\documentclass{article}
\usepackage{spconf,amsmath,graphicx}
\usepackage{amsfonts, url, ctable, tabularx,threeparttable}
\usepackage{amssymb}
\usepackage{pifont}
\newcommand{\cmark}{\ding{51}}%
\newcommand{\xmark}{\ding{55}}%

\newcolumntype{Y}{>{\centering\arraybackslash}X}

\title{3d human motion generation from the text via gesture action classification and the autoregressive model}
\name{Gwantae Kim$^{1}$, Youngsuk Ryu$^{1}$, Junyeop Lee$^{1}$ David K. Han$^2$, Jeongmin Bae$^{3}$ and Hanseok Ko$^{1}$}
\address{$^{1}$School of Electrical Engineering, Korea University, Seoul, South Korea\\
$^2$Drexel University,Philadelphia, PA, USA\\
$^{3}$DMLab, Seoul, South Korea}
\begin{document}
\ninept
\maketitle
\begin{abstract}
In this paper, a deep learning-based model for 3D human motion generation from the text is proposed via gesture action classification and an autoregressive model. The model focuses on generating special gestures that express human thinking, such as waving and nodding. To achieve the goal, the proposed method predicts expression from the sentences using a text classification model based on a pretrained language model and generates gestures using the gate recurrent unit-based autoregressive model. Especially, we proposed the loss for the embedding space for restoring raw motions and generating intermediate motions well. Moreover, the novel data augmentation method and stop token are proposed to generate variable length motions. To evaluate the text classification model and 3D human motion generation model, a gesture action classification dataset and action-based gesture dataset are collected. With several experiments, the proposed method successfully generates perceptually natural and realistic 3D human motion from the text. Moreover, we verified the effectiveness of the proposed method using a public-available action recognition dataset to evaluate cross-dataset generalization performance.
\end{abstract}
\section{Introduction}
\label{sec:intro}
{\let\thefootnote\relax\footnotetext{The authors of Korea University were supported by DMLab (Q2109331). Corresponding Author:Hanseok Ko.
}}
Human-like agents, including virtual avatars and social robots, are required to act with appropriate speech and accompanying gestures when interacting with humans. Recently, Pretrained Language Model(PLM)\cite{devlin2018bert, brown2020language} and speech synthesis\cite{wang2017tacotron, shen2018natural, ren2020fastspeech} have emerged as powerful instruments for text and speech generation. However, research on gesture generation as an alternate means of communicating with humans hasn't been as active.

Yoon et al.\cite{yoon2020speech} proposed a model for gesture generation from speech and text specifically tailored for the speaker. While the proposed model can generate gestures from an arbitrary context, the generated gestures were mostly neutral and the model wasn't able to produce specific gestures for delivering expressions, such as wave, handshake, and nod. This can be attributed in part due to the lack of datasets containing a large variety of gestures.

Clustering and continuity in the embedding space of gestures are needed for generating meaningful gestures with smooth inter-gesture transitions. However, previous works haven't been successful in addressing this issue. For example, Yu et al.\cite{yu2020structure} take random noise as input and action label, and their model performed reasonably for most of the gesture categories. However, it can only control some variations of certain actions and it lacks smooth inter-gesture transitions. GlocalNet\cite{battan2021glocalnet} tried to control motions in the embedding space. The distance between the action labels in the embedding space, however, turned out to be insufficient.  Thus, the model generated ambiguous motions for certain action labels. When the random seed was placed in certain parts of the embedding space, resultant gestures have been completely nonsensical. This is stemming from discontinuities in the embedding space. In order to ensure continuity among the generated gestures, we developed a novel loss function for inter-class distance and intermediate embedding. 

Most of the methods\cite{guo2020action2motion, yu2020structure} for gesture generation is constrained with a fixed length of generated frames. Since gestures are variable in length, such an assumption may sometimes result in unnatural gestures. To combat this issue, we propose a novel data augmentation method and stop-token estimation to generate variable-length motions.

While there exist some text classification and gesture datasets, the dataset that addresses 3D motion generation from text is nonexistent to the best of our knowledge.  For the text-to-gesture task, using existing datasets intended for text or gesture classification is quite challenging since it would result in much noise and manual labeling. For creating a natural human-like avatar with the ability to generate a variety of gestures accompanying the speech, we developed and present a dataset consisting of 12 gesture categories labeled in 1200 sample video clips.

\begin{figure}[t] 
\begin{center}
\includegraphics[width=0.6\linewidth]{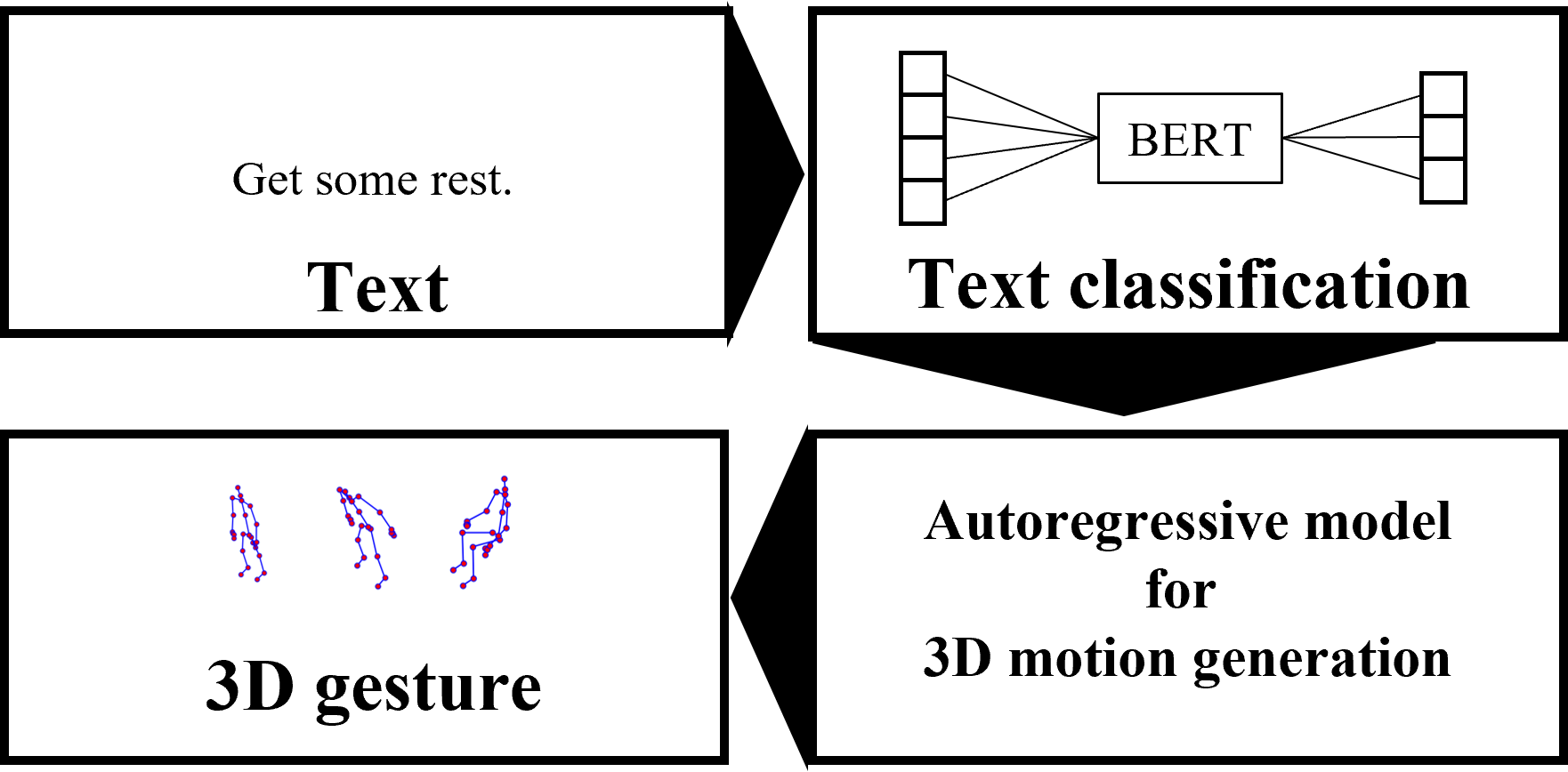}
\end{center}
\caption{Pipeline of the proposed method}
\label{fig : pipeline}
\end{figure}

The pipeline of the proposed method is shown in Fig.1. We use BERT\cite{devlin2018bert} PLM as a multitasking text classification model to find states from the text. Next, the motion generation model estimates the 3D motion sequence from the action class label. We developed a novel embedding loss for generating raw motions and intermediate motions from the latent vector. Additionally, we propose a novel data augmentation method and stop-token estimation to generate variable-length motions.

Our contributions are:
\begin{itemize}
    \item We propose a 3D human motion generation method from text using PLM and an autoregressive model for generating a large variety of gestures.
    \item We propose a novel data augmentation method and stop-token estimation to generate variable-length gesture motions.
    \item We developed a novel loss function for inter-class distance and intermediate embedding for ensuring continuity and smooth transition among the generated gestures.
    \item We developed a new text-to-gesture dataset consisting of 12 gesture categories labeled in 1200 sample video clips.
    
\end{itemize}

\begin{figure}[t] 
\begin{center}
\includegraphics[width=1.0\linewidth]{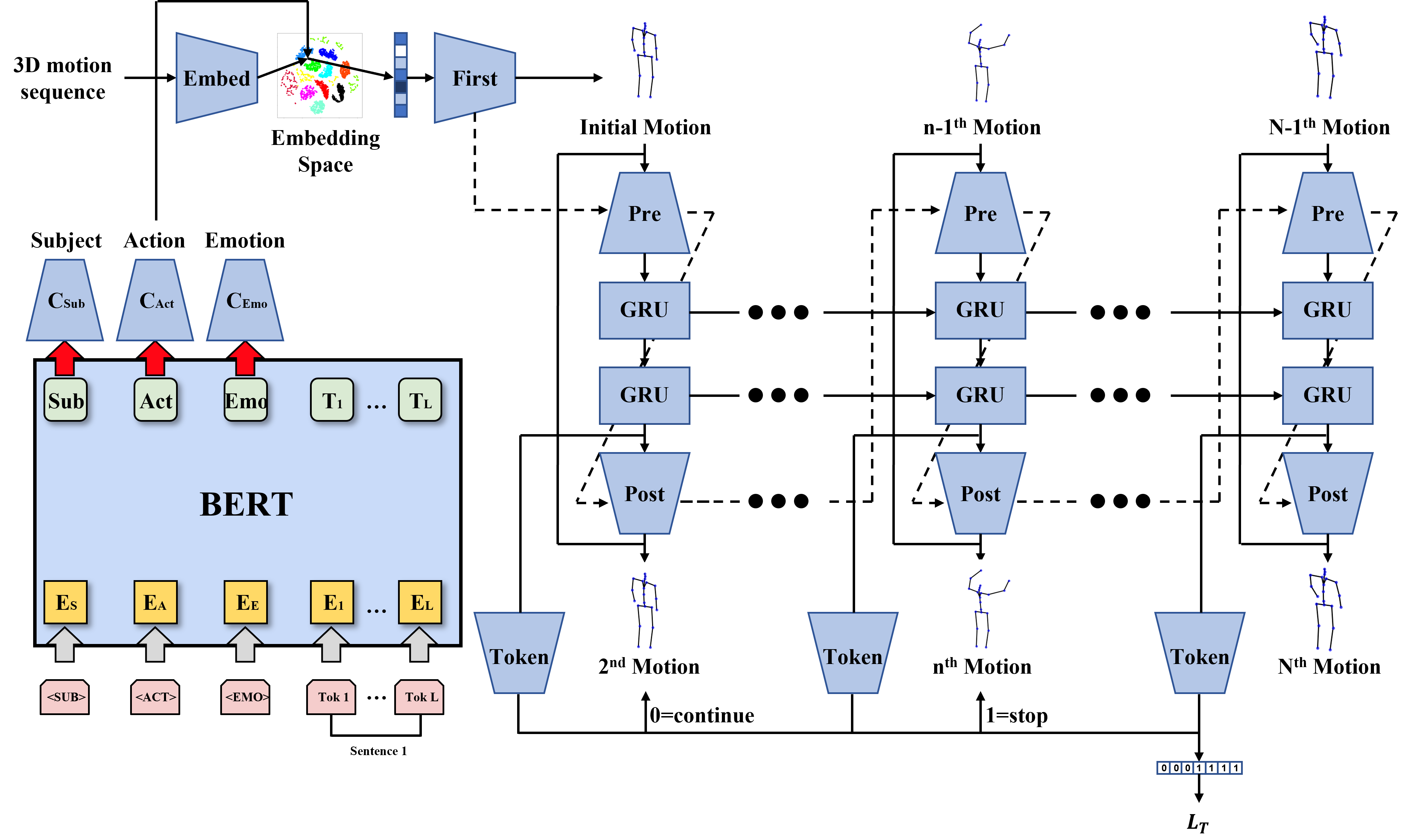}
\end{center}
\caption{The proposed 3D human motion generation framework from the text. (Left) Multitask text classification stage. (Right) 3D human motion generation stage.}
\label{fig : structure}
\end{figure}

\section{Related works}

\subsection{Pretrained language model}

An important element in our pipeline is to infer accompanying gestures from speech or text. PLMs are a powerful tool for such purpose since it models correlation between words and improves performance of downstream tasks, such as text classification\cite{minaee2021deep}, emotion recognition\cite{lee2021multimodal}, and intent classification\cite{chen2019bert}. There have been efforts of creating large-scale PLM, such as BERT\cite{devlin2018bert}, ALBERT\cite{lan2019albert}, and GPT-3\cite{brown2020language}. We take advantage of the existing large-scale PLMs to train our multitask text classification model.

\subsection{3D human motion generation}
In terms of generating 3D gestures, there have been some previous work-related parts to the model we propose here.
Martinez et al.\cite{martinez2017human} and Yan et al. \cite{yan2019convolutional} proposed models estimating future motion sequences from previous motion sequences, however, their performance drops when prior motion is ambiguous. Yoon et al.\cite{yoon2020speech} proposed gesture generation from speech. The generated gestures were simple ones and limited to a small number of classes since the needed text-gesture paired datasets are sparse. Some efforts have been devoted to estimating the whole motion sequence from the action labels\cite{cai2018deep, yu2020structure, guo2020action2motion}. Cai et al.\cite{cai2018deep} proposed two stage Generative Adversarial Network(GAN) with fully-connected layers. However, the model is hard to converge and cannot capture time-variant features. TimeGAN \cite{yoon2019time} proposed generative model for time series data. The model could train with various time-series dataset, but the generation performance is limited.

\section{proposed method}
\label{sec:proposed}
The structure of the proposed method is shown in Fig. 2. The proposed method is divided into two different stages: gesture action classification and 3D human motion generation. In the gesture action classification stage, the state of the sentence is predicted by a PLM-based fine-tuned model. If the sentence is classified such that it may accompany a gesture, then the 3D human motion generation model makes a motion sequence that is most appropriate for the sentence.

\subsection{Gesture Action Classification}
It is impossible to learn the distance between arbitrary words from scratch with a small amount of data. Therefore, the classification model is trained by fine-tuning a PLM that has learned the relationship between words in advance. We used the BERT\cite{devlin2018bert} model as the PLM. To solve the present problem, we constructed a model that classifies subject, action, and emotion at the same time. The classification models, $C_{sub}$, $C_{act}$, and $C_{emo}$, consist of two Fully Connected(FC) layers, each with hidden layer size=256, with ReLU activation, and Dropout p=0.3. While our model only makes use of action classification for the subsequent process, correct classification of subject and emotion has been shown to help improve the accuracy of the gesture classification.

The cross-entropy loss is used to train the gesture action classification model. Let x be an input sentence, then the classification loss is calculated by
\begin{equation}
    \mathcal{L}_{sub} = \mathcal{L}_{CE}(C_{sub}(BERT(x)), s)
\end{equation}
\begin{equation}
    \mathcal{L}_{act} = \mathcal{L}_{CE}(C_{act}(BERT(x)), a)
\end{equation}
\begin{equation}
    \mathcal{L}_{emo} = \mathcal{L}_{CE}(C_{emo}(BERT(x)), e)
\end{equation}
\begin{equation}
    \mathcal{L} = \mathcal{L}_{sub} + \mathcal{L}_{act} + \mathcal{L}_{emo}
\end{equation}
where s, a, and e are subject, action, and emotion labels, respectively. Adam optimizer\cite{kingma2014adam} is used for model training, where mini-batch size is 32, the epoch is 100, and the learning rate is 1e-5.

\subsection{3D human motion generation}
The 3D motion generation model consists of an embedder and a decoder. The embedder uses a labeled set of 3D motion sequences and is trained by deep metric learning\cite{schroff2015facenet} in supervised learning to estimate latent space that can classify action classes. The decoder is constructed using a Gated Recurrent Units(GRU) based autoregressive model to generate a sequence of motions as shown in Figure 2. To preserve information of the previous frames in long gesture sequences, skip connections are used.

\noindent\textbf{Data augmentation} Since frame lengths of raw motion data are not generally fixed, the frame length must be fixed using padding or cropping. Zero padding is frequently used\cite{wang2020learning, yu2020structure}, but the generated motion using zero-padded data often results in all the joints converging to zero. To solve the problem, our proposed padding strategy is to identically copy the last motion frame's joint coordinates to the remaining frames, essentially freezing the joints on the rest of the frames. To ensure that the generator is trained to stop generating the motion after a gesture is completed, we embed a token of 0 from the first to the last of the motion frames and embed 1 for the frozen motion frames.
The augmentation method applied here is to randomly select some of the motion frames and removed them from the raw data to generate one augmented motion clip. The process is repeated ten times over the same clip to generate ten slightly different augmented clips.

\noindent\textbf{Model structure}
The input of the embedder is 3D motion sequence data, The core element of the embedder consists of a Batch Normalization(BN) layer, a ReLU activation and a Convolution layer, and these are repeated five times with the convolution layer parameters of channel, kernel, and stride as follows: (64, 7, 2), (128, 3, 2), (256, 3, 2), (512, 3, 1), (1024, 3, 1). After applying global average pooling, the output is flattened to a 1024 dimensional vector spanning the embedding space.

The 1024 dimensional embedding vector is fed to the initial motion generator which consists of 4 FC layers with LeakyReLU activation with the hidden size of 512. The output of the second FC layer becomes an initial hidden state of GRU in the pre-net. After the initial motion is generated, it passes through pre-net, 2 GRU layers with hidden size 1024, and post-net. The pre-net and post-net have the same structure, each consisting of 2 FC layers with LeakyReLU activation, one GRU layer, and 2 FC layers with LeakyReLU activation. The size of every hidden layer is 512. Moreover, the residual connection is applied to prevent any unnaturally sudden change of motion. It is designed to share the GRU hidden state of the pre-net and the post-net to enhance information transfer and long-term dependency of the model(see dotted line in Fig. 2).

The token net consists of 3 FC layers with hidden size 512 and LeakyReLU activation. The output is scalar and a sigmoid function is applied to the output of the token net.

The model always generates the motion and token to maximum frame length N in the training phase, and the model generates the motion and token until the rounded output of the token net is being 1 in the test phase. Thus, the model can generate variable frame length data by the stop token.

\noindent\textbf{Training details}
The embedder E is pretrained by the Triplet loss\cite{schroff2015facenet}. Let \textbf{f} be a 3D motion sequence, then the loss is calculated by

\begin{equation}
    \mathcal{L}_{E} = [\mathcal{L}_{1}(E(\mathbf{f})_{anc}, E(\mathbf{f})_{pos}) - \mathcal{L}_{1}(E(\mathbf{f})_{anc}, E(\mathbf{f})_{neg}) + 0.05]_{+}
\end{equation}
where anc, pos, and neg denote anchor, positive, and negative samples, respectively. The triplet is sampled by all possible triplets in the mini-batch.

The L1 loss is used to train the generator model G and token model T with a 3D motion sequence and token. Since the teacher-forcing is not applied to train the model, generating a good initial frame is important to generate the whole motion. Therefore, we add more weight to generate initial frame generation with $\mathcal{L_F}$ loss. Let \textbf{v} be an embedding vector and \textbf{t} be a token, then the reconstruction loss for the first frame ($\mathcal{L}_F$) is calculated by
\begin{equation}
    \mathcal{L}_{F} = \mathcal{L}_{1}(G(\mathbf{v})[0], \mathbf{f}[0])
\end{equation}
To prevent the model generating still motion, the generator loss is not calculated for padded frames. The remaining frames reconstruction loss ($\mathcal{L}_G$) and the token loss ($\mathcal{L}_{T}$) are computed as
\begin{equation}
    \mathcal{L}_{G} = \sum_{n=0}^{L}\mathcal{L}_{1}(G(\mathbf{v})[n], \mathbf{f}[n])
\end{equation}
\begin{equation}
    \mathcal{L}_{T} = \sum_{n=1}^{N}\mathcal{L}_{1}(T(G(\mathbf{v}))[n], \mathbf{t}[n])
\end{equation}
where L is the frame length of the raw data. After the token loss converges, the model can estimate motion length by the token.  Moreover, to enhance the relationship between embedding vector and fake motion, we calculate a fake embedding vector loss.
\begin{equation}
    \mathcal{L}_{E} = \mathcal{L}_{1}(E(G(\mathbf{v})), \mathbf{v})
\end{equation}
The final loss is calculated by
\begin{equation}
    \mathcal{L}=
        \begin{cases}
            \mathcal{L}_{F} + \mathcal{L}_{G} + \mathcal{L}_{T} & \text{if $\mathcal{L}_T > 0.01$} \\
            \mathcal{L}_{F} + \mathcal{L}_{G} + \mathcal{L}_{E} & \text{else}
        \end{cases}
\end{equation}
Adam optimizer\cite{kingma2014adam} is used for model training, where mini-batch size is 32, the epoch is 500, and the learning rate is 1e-4. 

\begin{figure}[t] 
\begin{center}
\includegraphics[width=0.9\linewidth]{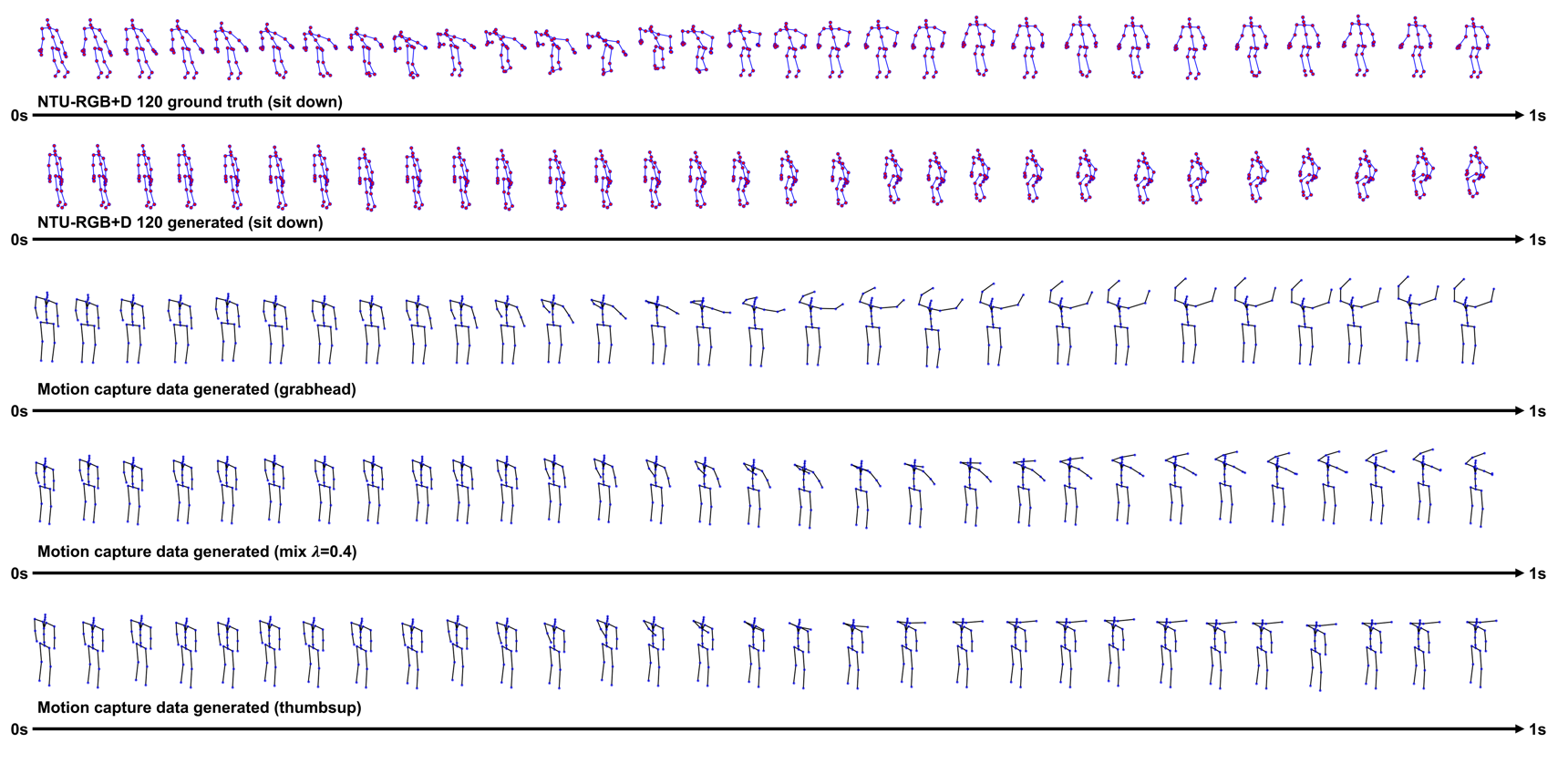}
\end{center}
\caption{Randomly-selected generation Results. The first row is ground-truth motion and the second row is generated motion using NTU-RGB+D 120 dataset. The third, fourth, fifth row are generated motion using motion capture dataset.}
\label{fig : result}
\end{figure}

\section{Experiments}
\label{sec:ex}
\subsection{Datasets}
\noindent\textbf{Gesture Action Classification} There are many datasets available for finding expressions from text (e.g., sentiment classification\cite{blitzer2007biographies} and intent classification\cite{larson2019evaluation}), but their label is not suitable to generate motion sequence. Thus, we developed a new dataset of text-to-gesture. The dataset contains 1050 training samples and 259 test samples, which are collected from the transcription of Youtube conversation videos. Each video clip with speech is turned into a sentence with the subject, the gesture, and the emotion labeled. The subject label has 16 classes: \textit{greeting, gratitude, farewell, information, inquiry, apology, praise, anger, congratulation, sympathy, cheer, concern, request, surprise, complaint, neutral}. The gesture label has 12 classes: \textit{handshake, point, thumbsup, nod, shakehead, wave, grabhead, pat, uplift, yay, shrug, neutral}. The emotion label has 3 classes: \textit{positive, negative, neutral}. Only the gesture label is applied in the model.

\noindent\textbf{3D human motion generation} To generate the 3D human motion from the gesture label, extracted from the gesture action classifier, we developed a new dataset, which consists of (gesture label, 3D joint trajectory) pair, using a motion capture device. In the dataset, 100 samples were collected for each gesture label, and a total of 1200 samples were collected. Each sample has 1 to 3 seconds long frames captured at 60fps, each frame has 21 joints, and each joint has a 3D joint location. The dataset is separated to 95\% as train set and 5\% as test set.

\noindent\textbf{NTU-RGB+D 120}
To evaluate the cross-dataset generalization performance of the proposed 3D human motion generation model, we used NTU-RGB+D 120\cite{liu2019ntu} dataset. Its 3D motion data is from Microsoft Kinect readout, which is known to be unreliable and temporally inconsistent\cite{guo2020action2motion}. Thus, NTU-RGB+D 120 dataset is a more challenging dataset compared to the dataset using a motion capture device. NTU-RGB+D 120 dataset contains 120 action classes, therefore utilizing the entire dataset is not appropriate for our model which is designed to generate motions in the order of a dozen classes.  To make the problem compatible, we have selected 12 classes, which are \textit{sit down, clapping, cheer up, hand waving, jump up, head nodding, head shaking, kicking, finger-pointing, hugging, walking apart}, and \textit{thumb up}. We apply \textit{cross-subject} setting\cite{liu2019ntu} for train/test data split.

\subsection{Evaluation metrics}
\noindent\textbf{Maximum Mean Discrepancy} The Maximum Mean Discrepancy(MMD) metric is based on a two-sample text to measure the similarity between two distributions. It is widely used to measure the quality of generated samples compared with real data in deep generative model\cite{zhao2019self} and human action generation\cite{yu2020structure}. The metric has also been applied to evaluate the similarity between generated actions and the ground truth in \cite{wang2020learning}, which has been proved consistent with human evaluation. Following the settings from \cite{yu2020structure}, we used $MMD_{avg}$ and $MMD_{seq}$ for evaluation.

\noindent\textbf{Recognition accuracy} We pre-train a recognition network on the training data to compute the classification accuracy of generated samples. We adopted EfficientGCN\cite{song2021constructing} model, which is skeleton-based action recognition network. These evaluation metrics can examine whether the conditional generated samples are actually residing in the same manifold as the ground truth and can be correctly recognized.

\subsection{Gesture action classification}
We evaluated the effectiveness of multitask learning in action label classification from the text. The backbone network is BERT\cite{devlin2018bert}. The experimental results are shown in Table 1. The result shows that the model achieved the best action classification performance when estimating the subject, action, and emotion labels at once. Since the subject and emotion are not independent of the action in the sentence, the subject classification, and emotion classification help to improve action classification.

\begin{table}[t]
\caption{Experimental results on multitask text classification dataset}
\label{table:ex1}
\centering
    \resizebox{\linewidth}{!}{
    \begin{tabularx}{\linewidth}{|Y||Y||Y||Y||Y||Y||Y||Y||Y|}
    \specialrule{.2em}{.1em}{.1em}
    \multicolumn{3}{c|}{label} & 
    \multicolumn{6}{c}{$Accuracy(\%)\uparrow$} \\ \hline
    \multicolumn{1}{Y|}{Sub.} & 
    \multicolumn{1}{Y|}{Act.} &
    \multicolumn{1}{Y|}{Emo.} & 
    \multicolumn{2}{Y|}{Sub.} &
    \multicolumn{2}{Y|}{Act.} &
    \multicolumn{2}{Y}{Emo.} \\ \hline
    \multicolumn{1}{Y|}{\xmark} & 
    \multicolumn{1}{Y|}{\cmark} &
    \multicolumn{1}{Y|}{\xmark} & 
    \multicolumn{2}{Y|}{-} &
    \multicolumn{2}{Y|}{46.12} &
    \multicolumn{2}{Y}{-} \\
    \multicolumn{1}{Y|}{\cmark} & 
    \multicolumn{1}{Y|}{\cmark} &
    \multicolumn{1}{Y|}{\xmark} & 
    \multicolumn{2}{Y|}{49.22} &
    \multicolumn{2}{Y|}{48.8} &
    \multicolumn{2}{Y}{-} \\
    \multicolumn{1}{Y|}{\xmark} & 
    \multicolumn{1}{Y|}{\cmark} &
    \multicolumn{1}{Y|}{\cmark} & 
    \multicolumn{2}{Y|}{-} &
    \multicolumn{2}{Y|}{50.00} &
    \multicolumn{2}{Y}{81.01} \\
    \multicolumn{1}{Y|}{\cmark} & 
    \multicolumn{1}{Y|}{\cmark} &
    \multicolumn{1}{Y|}{\cmark} & 
    \multicolumn{2}{Y|}{47.29} &
    \multicolumn{2}{Y|}{\textbf{50.77}} &
    \multicolumn{2}{Y}{81.78} \\
    \specialrule{.2em}{.1em}{.1em}
    \end{tabularx}}
\end{table}

\subsection{Quantitative results}
Evaluation results on motion capture and NTU-RGB+D 120 datasets are presented in Tables 2 and 3, respectively. Our approach outperforms the comparison methods on $MMD_{avg}$, $MMD_{seq}$, and accuracy on motion capture and NTU-RGB+D 120 datasets. Two-stage GAN\cite{cai2018deep} is hard to converge due to GAN loss, and TimeGAN\cite{yoon2019time} needs to train one model per one label. For recognition accuracy, the proposed model yields the highest accuracy, which suggests the potential to generate highly recognizable motions and to capture the characteristics of action types. We also evaluate the model on the frame length difference. The proposed approach achieved 16.79 mean frame length error on the motion capture dataset and 14.5 mean frame length error on NTU-RGB+D 120 dataset. Therefore, compared to the generator with fixed frame length, the proposed model can generate variable length motions with small errors.
\begin{table}[t]
\caption{Model comparisons in terms of MMD on motion capture dataset.}
\label{table:ex2}
\centering
    \resizebox{\linewidth}{!}{
    \begin{tabularx}{\linewidth}{|Y||Y||Y||Y|}
    \specialrule{.2em}{.1em}{.1em}
    \multicolumn{1}{Y|}{Models} & 
    \multicolumn{1}{Y|}{$MMD_{avg}\downarrow$} &
    \multicolumn{1}{Y|}{$MMD_{seq}\downarrow$} & 
    \multicolumn{1}{Y}{$Acc(\%)\uparrow$}\\ \hline
    \multicolumn{1}{Y|}{\cite{cai2018deep}} & 
    \multicolumn{1}{Y|}{1.310} &
    \multicolumn{1}{Y|}{1.233} & 
    \multicolumn{1}{Y}{8.30} \\
    \multicolumn{1}{Y|}{\cite{yoon2019time}} & 
    \multicolumn{1}{Y|}{1.172} &
    \multicolumn{1}{Y|}{1.188} & 
    \multicolumn{1}{Y}{41.67} \\
    \multicolumn{1}{Y|}{Proposed} & 
    \multicolumn{1}{Y|}{\textbf{0.223}} &
    \multicolumn{1}{Y|}{\textbf{0.230}} & 
    \multicolumn{1}{Y}{\textbf{77.80}} \\
    \multicolumn{1}{Y|}{GT} & 
    \multicolumn{1}{Y|}{0} &
    \multicolumn{1}{Y|}{0} & 
    \multicolumn{1}{Y}{100.00} \\
    \specialrule{.2em}{.1em}{.1em}
    \end{tabularx}}
\end{table}

\begin{table}[t]
\caption{Model comparisons in terms of MMD on NTU-RGB+D 120.}
\label{table:ex3}
\centering
    \resizebox{\linewidth}{!}{
    \begin{tabularx}{\linewidth}{|Y||Y||Y||Y||Y|}
    \specialrule{.2em}{.1em}{.1em}
    \multicolumn{1}{Y|}{Models} & 
    \multicolumn{1}{Y|}{$MMD_{avg}\downarrow$} &
    \multicolumn{1}{Y|}{$MMD_{seq}\downarrow$} & 
    \multicolumn{1}{Y}{$Acc(\%)\uparrow$}\\ \hline
    \multicolumn{1}{Y|}{\cite{cai2018deep}} & 
    \multicolumn{1}{Y|}{0.590} &
    \multicolumn{1}{Y|}{1.258} & 
    \multicolumn{1}{Y}{7.6} \\
    \multicolumn{1}{Y|}{\cite{yoon2019time}} & 
    \multicolumn{1}{Y|}{0.910} &
    \multicolumn{1}{Y|}{0.880} & 
    \multicolumn{1}{Y}{18.2} \\
    \multicolumn{1}{Y|}{Proposed} & 
    \multicolumn{1}{Y|}{\textbf{0.311}} &
    \multicolumn{1}{Y|}{\textbf{0.294}} & 
    \multicolumn{1}{Y}{\textbf{22.1}} \\
    \multicolumn{1}{Y|}{GT} & 
    \multicolumn{1}{Y|}{0} &
    \multicolumn{1}{Y|}{0} & 
    \multicolumn{1}{Y}{82.5} \\
    \specialrule{.2em}{.1em}{.1em}
    \end{tabularx}}
\end{table}

\subsection{Qualitative results}
We present some generated actions in the motion capture dataset and NTU RGB+D 120 dataset in Fig. 3. The first row is ground-truth motion and the second row is generated by the proposed model. The action label is \textit{sit down}. The generated motion is not the same as the original motion, but it is perceptually natural and realistic. The next three-row are generated motions by the proposed model. The embedding vector of the third row is the mean vector of the test set in \textit{grabhead} label and the embedding vector of the fifth row is the mean vector of the test set in \textit{thumbsup} label. Both generated good results. The embedding vector of the fourth row is an intermediate vector that is the weighted sum vector between the mean vector for \textit{grabhead} and the mean vector for \textit{thumbsup} label. The generated result using the intermediate embedding vector successfully mixed the motions of the two different labels. We present the software that generates a 3D motion sequence from the text here\footnote{\url{https://github.com/GT-KIM/motion_generation_from_text}}.

\section{conclusion}
\label{sec:conclusion}
We presented a novel deep learning-based method for 3D human motion generation from the text using PLM and an autoregressive model. We collected two new datasets for multitasking text state classification and 3D human motion generation tasks. Through a series of experiments, the proposed method successfully showed that it generates the 3D human motion sequence from the text. We also evaluated the cross-dataset generalization performance of the proposed generation model using the NTU-RGB+D 120 dataset. The proposed method can be extended to generate the motions for the avatar-based interactive agents and service robots. In our future work, we plan to study a model that directly generates 3D human motion from text without an action class.

\bibliographystyle{IEEEbib}
\bibliography{ref}

\end{document}